\documentclass[11pt]{article}
\usepackage[utf8]{inputenc}
\usepackage{eacl2017}
\usepackage{times}
\usepackage{url}
\usepackage{latexsym}
\usepackage{booktabs}
\usepackage{tikz-dependency}
\usepackage{graphicx}
\usepackage{float}
\usepackage{multirow}

\eaclfinalcopy
\def\eaclpaperid{173}

\makeatletter
\newcommand{\@BIBLABEL}{\@emptybiblabel}
\newcommand{\@emptybiblabel}[1]{}
\makeatother

\title{Distant Supervision for Relation Extraction beyond the Sentence Boundary}

\author{Chris Quirk \and Hoifung Poon \\
  Microsoft Research\\
  One Microsoft Way\\
  Redmond, WA 98052 \\
  {\tt \{chrisq,hoifung\}@microsoft.com} 
  \\}

\begin{document}
\maketitle

\begin{abstract}
The growing demand for structured knowledge has led to great interest in relation extraction, especially in cases with limited supervision.
However, existing distance supervision approaches only extract relations expressed in single sentences.
In general, cross-sentence relation extraction is under-explored, even in the supervised-learning setting.
In this paper, we propose the first approach for applying distant supervision to cross-sentence relation extraction.
At the core of our approach is a graph representation that can incorporate both standard dependencies and discourse relations, thus providing a unifying way to model relations within and across sentences.
We extract features from multiple paths in this graph, increasing accuracy and robustness when confronted with linguistic variation and analysis error.
Experiments on an important extraction task for precision medicine show that our approach can learn an accurate cross-sentence extractor, using only a small existing knowledge base and unlabeled text from biomedical research articles.
Compared to the existing distant supervision paradigm, our approach extracted twice as many relations at similar precision, thus demonstrating the prevalence of cross-sentence relations and the promise of our approach.
\end{abstract}

\section{Introduction}

The accelerating pace in technological advance and scientific discovery has led to an explosive growth in knowledge. The ensuing information overload creates new urgency in assimilating fragmented knowledge for integration and reasoning.
A salient case in point is precision medicine \cite{bahcall15}. 
The cost of sequencing a person's genome has fallen below \$1000\footnote{\url{http://www.illumina.com/systems/hiseq-x-sequencing-system.html}}, enabling individualized diagnosis and treatment of complex genetic diseases such as cancer. 
The availability of measurement for 20,000 human genes makes it imperative to integrate all knowledge about them, which grows rapidly and is scattered in millions of articles in PubMed\footnote{\url{http://www.ncbi.nlm.nih.gov/pubmed}}. 
Traditional extraction approaches require annotated examples, which makes it difficult to scale to the explosion of extraction demands.
Consequently, there has been increasing interest in indirect supervision \cite{banko&al07,poon&domingos09,toutanova&al15}, with distant supervision \cite{craven&al98,mintz&al09} emerging as a particularly promising paradigm for augmenting existing knowledge bases from unlabeled text \cite{poon&al15,parikh&al15}.

This progress is exciting, but distant-supervision approaches have so far been limited to single sentences,
thus missing out on relations crossing the sentence boundary.
Consider the following example:{\small \em ``The p56Lck inhibitor {\bf Dasatinib} was shown to enhance apoptosis induction by dexamethasone in otherwise GC-resistant CLL cells. 
This finding concurs with the observation by Sade showing that {\bf Notch}-mediated resistance of a mouse lymphoma cell line could be overcome by inhibiting p56Lck.''}
Together, the two sentences convey the fact that the drug {\em Dasatinib} could overcome resistance conferred by mutations to the {\em Notch} gene, which can not be inferred from either sentence alone.
The impact of missed opportunities is especially pronounced in the long tail of knowledge. Such information is crucial for integrative reasoning as it includes the newest findings in specialized domains.

In this paper, we present {\em DISCREX}, the first approach for distant supervision to relation extraction beyond the sentence boundary. The key idea is to adopt a document-level graph representation that augments conventional intra-sentential dependencies with new dependencies introduced for adjacent sentences and discourse relations.
It provides a unifying way to derive features for classifying relations between entity pairs.
As we augment this graph with new arcs, the number of possible paths between entities grow.
We demonstrate that feature extraction along multiple paths leads to more robust extraction, allowing the learner to find structural patterns even when the language varies or the parser makes an error.

The cross-sentence scenario presents a new challenge in candidate selection.
This motivates our concept of {\em minimal-span candidates} in Section~\ref{sec:minimal}. Excluding non-minimal candidates substantially improves classification accuracy.

There is a long line of research on discourse phenomena, including coreference \cite{haghighi&klein07,poon&domingos08,rahman&ng09,raghunathan&al10}, narrative structures \cite{chambers&jurafsky09,cheung&al13}, and rhetorical relations \cite{marcu00}.
For the most part, this work has not been connected to relation extraction. 
Our proposed extraction framework makes it easy to integrate such discourse relations.
Our experiments evaluated the impact of coreference and discourse parsing, a preliminary step toward in-depth integration with discourse research.

We conducted experiments on extracting drug-gene interactions from biomedical literature, an important task for precision medicine.
By bootstrapping from a recently curated knowledge base (KB) with about 162 known interactions, our DISCREX system learned to extract inter-sentence drug-gene interactions at high precision. Cross-sentence extraction doubled the yield compared to single-sentence extraction. Overall, by applying distant supervision, we extracted about 64,000 distinct interactions from about one million PubMed Central full-text articles, attaining two orders of magnitude increase compared to the original KB.

\section{Related Work}

To the best of our knowledge, distant supervision has not been applied to cross-sentence relation extraction in the past.
For example, Mintz et al. \shortcite{mintz&al09}, who coined the term ``distant supervision'', aggregated features from multiple instances for the same relation triple (relation, entity1, entity2), but each instance is a sentence where the two entities co-occur. 
Thus their approach cannot extract relations where the two entities reside in different sentences. Similarly, Zheng et al. \shortcite{zheng&al16} aggregated information from multiple sentential instances, but could not extract cross-sentence relations.

Distant supervision has also been applied to completing Wikipedia Infoboxes \cite{wu&weld07} or TAC KBP Slot Filling\footnote{\url{http://www.nist.gov/tac/2016/KBP/ColdStart/index.html}}, where the goal is to extract attributes for a given entity, which could be considered a special kind of relation triples (attribute, entity, value). These scenarios are very different from general cross-sentence relation extraction. For example, the entity in consideration is often the protagonist in the document (title entity of the article). 
Moreover, state-of-the-art methods typically consider extracting from single sentences only \cite{surdeanu&al12,surdeanu&ji14,koch&al14}.

In general, cross-sentence relation extraction has received little attention, even in the supervised-learning setting.
Among the limited amount of prior work, Swampillai \& Stevenson \shortcite{swampillai&stevenson11} is the most relevant to our approach, as it also considered syntactic features and introduced a dependency link between the root nodes of parse trees containing the given pair of entities.
However, the differences are substantial. 
First and foremost, their approach used standard supervised learning rather than distant supervision. 
Moreover, we introduced the document-level graph representation, which is much more general, capable of incorporating a diverse set of discourse relations and enabling the use of rich syntactic and surface features (Section~\ref{sec:mthd}). 
Finally, Swampillai \& Stevenson \shortcite{swampillai&stevenson11} evaluated on MUC6\footnote{\url{https://catalog.ldc.upenn.edu/LDC2003T13}}, which contains only 318 Wall Street Journal articles. In contrast, we evaluated on large-scale extraction from about one million full-text articles and demonstrated the large impact of cross-sentence extraction for an important real-world application.

The lack of prior work in cross-sentence relation extraction may be partially explained by the domains of focus.
Prior extraction work focuses on newswire text\footnote{E.g., MUC6, ACE \url{https://www.ldc.upenn.edu/collaborations/past-projects/ace}} and the Web \cite{craven&al00}.
In these domains, the extracted relations often involve popular entities, for which there often exist single sentences expressing the relation \cite{banko&al07}.
However, there is much less redundancy in specialized domains such as the frontiers of science and technology, where cross-sentence extraction is more likely to have a significant impact.
The long-tailed characteristics of such domains also make distant supervision a natural choice for scaling up learning.
This paper represents a first step toward exploring the confluence of these two directions.

Distant supervision has been extended to capture implicit reasoning, via matrix factorization or knowledge base embedding \cite{riedel&al13,toutanova&al15,toutanova&al16}. Additionally, various models have been proposed to address the noise in distant supervision labels \cite{hoffmann&al11,surdeanu&al12}. 
These directions are orthogonal to cross-sentence extraction, and incorporating them will be interesting future work.

Recently, there has been increasing interest in relation extraction for biomedical applications \cite{kim&al09,nedellec&al13}. However, past methods are generally limited to single sentences, whether using supervised learning \cite{bjorne&al09,poon&vanderwende10,riedel&mccallum11} or distant supervision \cite{poon&al15,parikh&al15}. 

The idea of leveraging graph representations has been explored in many other settings, such as knowledge base completion \cite{lao&al11,gardner&mitchell15}, frame-semantic parsing \cite{das&smith11}, and other NLP tasks \cite{radev&mihalcea08,subramanya&al10}.
 Linear and dependency paths are popular features for relation extraction \cite{snow&al06,mintz&al09}. However, past extraction focuses on single sentences, and typically considers the shortest path only.
In contrast, we allow interleaving edges from dependency and word adjacency, and consider top $K$ paths rather than just the shortest one. This resulted in substantial accuracy gain (Section 4.5).

There has been prior work on leveraging coreference in relation extraction, often in the standard supervised setting \cite{hajishirzi&al13,durrett&klein14}, but also in distant supervision \cite{koch&al14,AugensteinAl16}. 
Notably, while Koch et al.~\shortcite{koch&al14} and Augenstein et al.~\shortcite{AugensteinAl16} still learned to extract from single sentences, they augmented mentions with coreferent expressions to include linked entities that might be in a different sentence.
We explored the potential of this approach in our experiments, but found that it had little impact in our domain, as it produced few additional candidates  beyond single sentences.
Recently, discourse parsing has received renewed interest \cite{ji&eisenstein14,feng&hirst14,surdeanu&al15}, and discourse information has been shown to improve performance in applications such as question answering \cite{sharp&al15}.
In this paper, we generated coreference relations using the state-of-the-art Stanford coreference systems \cite{lee&al11,recasens&al13,clark&manning15}, and generated rhetorical relations using the winning approach \cite{WangLan15} in the CoNLL-2015 Shared Task on Discourse Parsing.

\begin{figure*}[t!]
\footnotesize
\begin{dependency}[theme = simple]
   \begin{deptext}[column sep=0.1em]
     The \& p56Lck \& inhibitor \& {\bf Dasatinib} \& was \& shown \& to \& enhance \& apoptosis \& induction \& in \& otherwise \& GC-resistant \& CLL \& cells \\
   \end{deptext}
   \deproot{6}{ROOT}
   \depedge{3}{1}{DET}
   \depedge{3}{2}{NN}
   \depedge{6}{3}{NSUBJPASS}
   \depedge{3}{4}{ABBREV}
   \depedge{6}{5}{AUXPASS}
   \depedge{6}{8}{XCOMP}
   \depedge{10}{9}{NN}
   \depedge{8}{10}{DOBJ}
   \depedge{13}{12}{ADVMOD}
   \depedge{15}{13}{AMOD}
   \depedge{15}{14}{NN}
   \depedge{10}{15}{PREP\_IN}

   \node (prepre) [xshift=-6.5em,yshift=-0.5em] {};
   \node (nextsent) [xshift=-21em,yshift=-2.5em] {};
   
   \draw[->] (prepre) -- (nextsent) ;
   \node (nextsentlab) [xshift=-10em,yshift=-1.5em] {\tiny\textsc{NEXTSENT}};
   
   \begin{deptext}[column sep=-0.2em]
   \\
   \\
     \& \& \& \& \& \& \& \& \& \& \& \& \& \& \& \& \& \& \\
     \& \& \& \& \& \& \& \& \& \& \& \& \& \& \& \& \& \& \\
     \& \& \& \& \& \& \& \& \& \& \& \& \& \& \& \& \& \& \\
     \& \& \& \& \& \& \& \& \& \& \& \& \& \& \& \& \& \& \\
     This \& shows \& that \& {\bf Notch } \& -mediated \& resistance \& of \& a \& mouse \& lymphoma \& cell \& line \& could \& be \& overcome \& by \& inhibiting \& p56Lck \& . \\
   \end{deptext}
   
   \deproot[edge below]{2}{ROOT}
   \depedge[edge below]{2}{1}{DET}
   \depedge[edge below]{15}{3}{COMPLM}
   \depedge[edge below]{5}{4}{HYPHEN}
   \depedge[edge below]{6}{5}{AMOD}
   \depedge[edge below]{15}{6}{NSUBJPASS}
   \depedge[edge below]{12}{8}{DET}
   \depedge[edge below]{12}{9}{NN}
   \depedge[edge below]{12}{10}{NN}
   \depedge[edge below]{12}{11}{NN}
   \depedge[edge below]{6}{12}{PREP\_OF}
   \depedge[edge below]{15}{13}{AUX}
   \depedge[edge below]{15}{14}{AUXPASS}
   \depedge[edge below]{2}{15}{CCOMP}
   \depedge[edge below]{15}{17}{AGENT}
   \depedge[edge below]{17}{18}{DOBJ}

\end{dependency}
\label{doc-graph}
\vspace{-4em}
\caption{
    An example document graph for two sentences. Edges represent conventional intra-sentential dependencies, as well as connections between the roots of adjacent sentences (\textsc{NEXTSENT}).
    For simplicity, we omit edges between adjacent words or representing discourse relations. 
}
\end{figure*}

\section{Distant Supervision for Cross-Sentence Relation Extraction}
\label{sec:mthd}

In this section, we present DISCREX, short for {\em DIstant Supervision for Cross-sentence Relation EXraction}.
Similar to conventional approaches, DISCREX learns a classifier to predict the relation between two entities, given text spans where the entities co-occur.
Unlike most existing methods, however, DISCREX allows text spans comprising multiple sentences and explores potentially many paths between these entities.

\subsection{Distant Supervision}

Like prior approaches, DISCREX learns from an existing knowledge base (KB) and unlabeled text.
The KB contains known instances for the given relation.
In a preprocessing step, relevant entities are annotated within this text using available entity extraction tools.
Co-occurring entity pairs known to have the relation in the KB are chosen as positive examples. 
Under the assumption that related entities are relatively rare, we randomly sample co-occurring entity pairs not known to have the relation as negative examples.
To ensure a balanced training set, we always sampled roughly the same number of negative examples as positive ones.

\subsection{Minimal-Span Candidates}
\label{sec:minimal}

In standard distant supervision, co-occurring entity pairs with known relations are enlisted as candidates of positive training examples.
This is reasonable when the entity pairs are within single sentences.
In the cross-sentence scenario, however, this would risk introducing too many wrong examples.
Consider the following two sentences:
{\em Since amuvatinib inhibits {\bf KIT}, we validated MET kinase inhibition as the primary cause of cell death. Additionally, {\bf imatinib} is known to inhibit KIT.}
The mention of drug-gene pair {\em imatinib} and {\em KIT} (in bold) span two sentences, but the same pair also co-occur in the second sentence alone. 
In general, one might find co-occurring entity pairs in a large text span, where the same pairs also co-occur in a smaller text span that overlaps with the larger one.
In such cases, if there is a relation between the pair, mostly likely it is expressed in the smaller text span when the entities are closer to each other.

This motivates us to define that an co-occurring entity pair has the {\em minimal span} if there does not exist another overlapping co-occurrence of the same pair where the distance between the entity mentions is smaller. Here, the distance is measured in the number of consecutive sentences between the two entities.
Experimentally, we compared extraction with or without the restriction to minimal-span candidates, and show that the former led to much higher extraction accuracy.

\subsection{Document Graph}

To derive features for entity pairs both within and across sentences, DISCREX introduces a {\em document graph} with nodes representing words and edges representing intra- and inter-sentential relations such as dependency, adjacency, and discourse relations.
Figure~1 
shows an example document graph spanning two sentences. 
Each node is labeled with its lexical item, lemma, and part-of-speech.
We used a conventional set of intra-sentential edges: typed, collapsed Stanford dependencies derived from syntactic parses \cite{marneffe&al06}.
To mitigate parser errors, we also add edges between adjacent words.

As for inter-sentential edges, a simple but intuitive approach is to add an edge between the dependency roots of adjacent sentences: if we imagined that each sentence participated as a node in a type of discourse dependency tree, this represents a simple right-branching baseline.
To gather a finer grained representation of rhetorical structure,
we ran a state-of-the-art discourse parser~\cite{WangLan15} to identify discourse relations, which returned a set of labeled binary relations between spans of words.
We found the shortest path between any word in the first span and any word in the second span using only dependency and adjacent sentence edges, and added an edge labeled with the discourse relation between these two words.
Another source of potentially cross-sentence links comes from coreference.
We generated coreference relations using the Stanford Coreference systems (both statistical and deterministic) \cite{lee&al11,recasens&al13,clark&manning15}, and added edges from anaphora to their antecedents.

We also considered a special case of cross-sentence relation extraction by augmenting single-sentence candidates with coreference \cite{koch&al14,AugensteinAl16}. Namely, extraction is still conducted within single sentences, yet entity linking is extended to consider all coreference mentions for a relation argument. 
However, this did not produce significantly more candidates (2\% more for positive examples), most of which were not cross-sentence ones (only 1\%).

\subsection{Features}

Dependency paths have been established as a particularly effective source for relation extraction features \cite{mintz&al09}. 
DISCREX generalizes this idea by defining feature templates over paths in the document graph, which may contain interleaving edges of various types (dependency, word and sentence adjacency, discourse relation).
Dependency paths provide interpretable and generalizable features but are subject to parser error.
One error mitigation strategy is to add edges between adjacent words, allowing multiple paths between entities.

Feature extraction begins with a pair of entities in the document graph that potentially are connected by a relation.
We begin by finding a path between the entities of interest, and extract features from that path.

Over each such path, we explore a number of different features.
Below, we assume that each path is a sequence of nodes and edges $(n_1, e_1, n_2, \ldots, e_{L-1}, n_{L})$, with $n_1$ and $n_L$ replaced by special entity marker nodes.\footnote{
This prevents our method from memorizing the entities in the original knowledge base.
}

\paragraph{Whole path features} 
We extract four binary indicator features for each whole path, with nodes $n_i$ represented by their lexical item, lemma, part-of-speech tag, or nothing.
These act as high precision but low recall indicators of useful paths.

\paragraph{Path n-gram features}
A more robust and generalizable approach is to consider a sliding window along each path.
For each position $i$, we extract $n$-gram ($n=1-5$) features starting at each node ($n_i$, then 
$n_i \cdot e_i$ and so on until 
$n_i \cdot e_i \cdot n_{i+1} \cdot e_{i+1} \cdot n_{i+2}$)
and each edge
($e_i$
up to 
$e_i \cdot n_{i+1} \cdot e_{i+1} \cdot n_{i+2} \cdot e_{i+2}$).
Again, each node could be represented by its lexical item, lemma, or part of speech, leading to 27 feature templates.
We add three more feature templates using only edge labels ($e_i$; $e_i \cdot e_{i+1}$; and $e_i \cdot e_{i+1} \cdot e_{i+2}$) for a total of 30 feature templates.

\subsection{Multiple paths}

Most prior work has only looked at the single shortest path between two entities.
When authors use consistent lexical and syntactic constructions, and when the parser finds the correct parse, this approach works well.
Real data, however, is quite noisy.

One way to mitigate errors and be robust against noise is to consider multiple possible paths.
Given a document graph with arcs of multiple types, there are often multiple paths between nodes.
For instance, we might navigate from the gene to the drug using only syntactic arcs, or only adjacency arcs, or some combination of the two.
Considering such variations gives more opportunities to find commonalities between seemingly disparate language.

We explore varying the number of shortest paths, $N$, between the nodes in the document graph corresponding to the relevant entities.
By default, all edge types have an equal weight of 1, except edges between adjacent words.
Empirically, penalizing adjacency edges led to substantial benefits, though including adjacency arcs was important for benefits from multiple paths.
This suggests that the parser produces valuable information, but that we should have a back-off strategy for accommodating parser errors.

\subsection{Evaluation}

There is no gold annotated dataset in distant supervision, so evaluation typically resorts to two strategies. One strategy uses held-out samples from the training dataset, essentially treating the noisy annotation as gold standard. This has the advantage of being automatic, but could produce biased results due to false negatives (i.e., entity pairs not known to have the relation might actually have the relation).
Another strategy reports absolute recall (number of extractions from all unlabeled text), as well as estimated precision by manually annotating extraction samples from general text.
We conducted both types of evaluation in the experiments.

\begin{figure*}[t!]
  \includegraphics[trim={0 3cm 0 0},clip,keepaspectratio=true,width=0.9\textwidth]{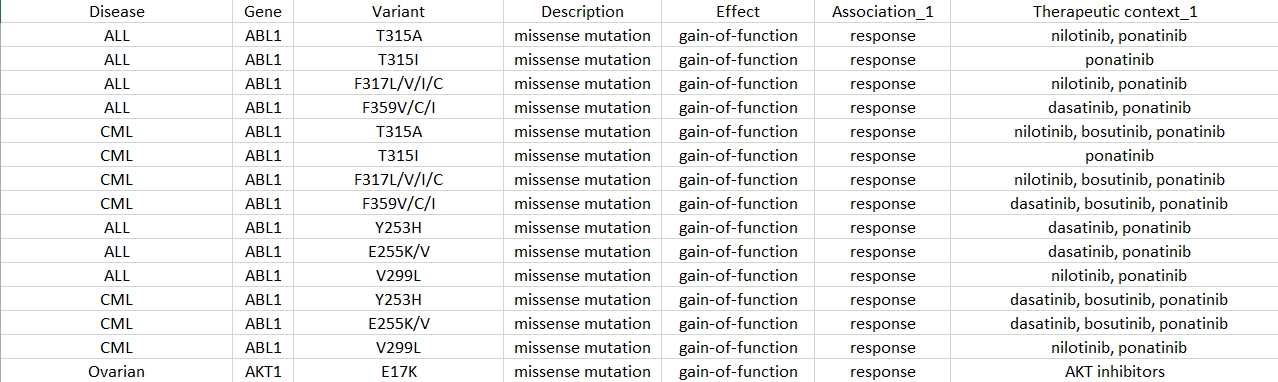}
  \vspace{-0.5em}
  \label{fg:gdkd}
  \caption{
  Sample rows from the Gene Drug Knowledge Database.
  Our current work focuses on two important columns: gene, and therapeutic context (drug).}
\end{figure*}

\section{Experiments}
\label{sec:exp}

We consider the task of extracting drug-gene interactions from biomedical literature.
A drug-gene interaction is broadly construed as an association between the drug efficacy and the gene status.
The status includes mutations and activity measurements (e.g., overexpression).
For simplicity, we only consider the relation at the drug-gene level, without distinguishing among details such as drug dosage or distinct gene status.

\subsection{Knowledge Base}

We used the Gene Drug Knowledge Database (GDKD) \cite{dienstmann&al15} for distant supervision. 
Figure~2 
shows a snapshot of the dataset.
Each row specifies a gene, some drugs, the fine-grained relations (e.g., sensitive), the gene status (e.g., mutation), and some supporting article IDs.
In this paper, we only consider the coarse drug-gene association and ignore the other fields.

\subsection{Unlabeled Text}

We obtained biomedical literature from PubMed Central\footnote{\url{http://www.ncbi.nlm.nih.gov/pmc/}}, which as of early 2015 contained about 960,000 full-text articles.
We preprocessed the text using SPLAT \cite{quirk&al12} to conduct tokenization, part-of-speech tagging, and syntactic parsing, and obtained Stanford dependencies \cite{marneffe&al06} using Stanford CoreNLP \cite{manning&al14}.
We used the entity taggers from Literome \cite{poon&al14} to identify drug and gene mentions.

\subsection{Candidate Selection}

\begin{table}[t]
    \centering
    \begin{tabular}{crr}
    \toprule
       \textbf{Number of Candidates}  &  {\bf $K=1$} & {\bf $K=3$} \\
    \midrule
        Unique Pairs & 169,168 & 332,969 \\
        Instances & 1,724,119 & 3,913,338 \\
        Matching GDKD & 58,523 & 87,773 \\
    \bottomrule
    \end{tabular}
    \caption{
        Statistics for drug-gene interaction candidates in PubMed Central articles: unique pairs, instances, instances with known relations in Gene Drug Knowledge Database (GDKD).
    }
    \label{tb:sliding}
\end{table}

To avoid unlikely candidates such as entity pairs far apart in the document, we consider entity pairs within $K$ consecutive sentences.
$K=1$ corresponds to extraction within single sentences.
For cross-sentence extraction, we chose $K=3$ as it doubled the number of overall candidates, while being reasonably small so as not to introduce too many unlikely ones.
Table~\ref{tb:sliding} shows the statistics of drug-gene interaction candidates identified in PubMed Central articles.
For $K=3$, there are 87,773 instances for which the drug-gene pair has known associations in Gene Drug Knowledge Database (GDKD), which are used as positive training examples.
Note that these only include minimal-span candidates (Section~\ref{sec:minimal}). Without the restriction, there are 225,520 instances matching GDKD, though many are likely false positives.

\subsection{Classifier}

Our classifiers were binary logistic regression models, trained to optimize log-likelihood with an $\ell_2$ regularizer.
We used a weight of 1 for the regularizer; the results were not very sensitive to the specific value.
Parameters were optimized using L-BFGS~\cite{nocedal&wright06}.
Rather than explicitly mapping each feature to its own dimension, we hashed the feature names and retained 22 bits~\cite{weinberger&al09}.
Approximately 4 million possible features seemed to suffice for our problem: fewer bits produced degradations, but more bits did not lead to improvements.

\subsection{Automatic Evaluation}

To evaluate the impact of features, we conducted five-fold cross validation, by treating the positive and negative examples from distant supervision as gold annotation. 
To avoid train-test contamination, all instances from a document are assigned to the same fold.
We then evaluated the average test performance across folds. 
Since our datasets were balanced by design (Section 3.1), we simply reported accuracy.
As discussed before, the results could be biased by the noise in annotation, but this automatic evaluation enables an efficient comparison of various design choices.

\begin{table}[t]
    \centering
    \begin{tabular}{lcc}
    \toprule
       \textbf{Features}  &  \textbf{Single-Sent.} & \textbf{Cross-Sent.} \\
    \midrule
        Base & 81.3 & 81.7 \\
    \midrule
        3 paths        & 85.4 & 85.5 \\
        ~~~+coref      & 85.0 & 84.7 \\
        ~~~+disc       & ---  & 84.6 \\
        ~~~+coref+disc & ---  & 84.5 \\
    \midrule
        10 paths       & 87.0 & 86.6 \\
        ~~~+coref      & 86.5 & 85.9 \\
        ~~~+disc       & ---  & 86.5 \\
        ~~~+coref+disc & ---  & 85.9 \\
    \bottomrule
    \end{tabular}
    \caption{
        Average test accuracy in five-fold cross-validation.
        Cross-sentence extraction was conducted within a sliding window of 3 sentences using minimal-span candidates.
        {\tt Base} only used the shortest path to construct features.
        \emph{3 paths} and \emph{10 paths} gathered features from the top three or ten shortest paths, assigning uniform weights to all edges except adjacency, which had a weight of 16.
        \emph{+coref} adds edges for the relations predicted by Stanford Coreference.
        \emph{+disc} adds edges for the predicted rhetorical relations by a state-of-the-art discourse parser \protect\cite{WangLan15}.
    }
    \label{tb:auto}
\end{table}

\begin{table}[t]
    \centering
    \begin{tabular}{cccc}
    \toprule
       \textbf{Paths} & \textbf{Adj. Wt.}&  \textbf{Single-Sent.} & \textbf{Cross-Sent.} \\
    \midrule
        \multirow{4}*{3}	& 1	& 82.2	& 82.1 \\
        	& 4	& 85.0	& 84.9 \\
        	& 16	& 85.4	& 85.5 \\
        	& 64	& 85.1	& 85.0 \\
    \midrule
        \multirow{4}*{10}	& 1	& 85.7	& 83.6 \\
        	& 4	& 87.2	& 86.7 \\
        	& 16	& 87.0	& 86.6 \\
        	& 64	& 87.0	& 86.6 \\
    \midrule
        \multirow{4}*{30}	& 1	& 87.6	& 85.4 \\
        	& 4	& 88.0	& 87.5 \\
        	& 16	& 87.5	& 87.2 \\
        	& 64	& 87.5	& 87.2 \\
    \bottomrule
    \end{tabular}
    \caption{
        Average test accuracy in five-fold cross-validation.
        Uniform weights are used, except for adjacent-word edges.
    }
    \label{tb:pathsWeights}
\end{table}

First, we set out to investigate the impact of edge types and path number.
We set the weight for adjacent-word edges to 16, to give higher priority to other edge types (weight 1) that are arguably more semantics-related.
Table~\ref{tb:auto} shows the average test accuracy for single-sentence and cross-sentence extraction with various edge types and path numbers.
Compared to extraction within single sentences, cross-sentence extraction attains a similar accuracy, even though the recall for the latter is much higher (Table 1).

Adding more paths other than the shortest one led to a substantial improvement in accuracy. The gain is consistent for both single-sentence and cross-sentence extraction.
This is surprising, as prior methods often derive features from the shortest dependency path alone.

Adding discourse relations, on the other hand, consistently led to a small drop in performance, especially when the path number is small.
Upon manual inspection, we found that Stanford Coreference made many errors in biomedical text, such as resolving a dummy pronoun with a nearby entity.
In hindsight, this is probably not surprising: state-of-the-art coreference systems are optimized for newswire domain and could be ill-suited for scientific literature \cite{bell&al16}.
We are less certain about why discourse parsing didn't seem to help. There are clearly examples where extraction errors could have been avoided given rhetorical relations (e.g., when the sentence containing the second entity starts a new topic). We leave more in-depth investigation to future work.

Next, we further evaluated the impact of path number and adjacency edge weight.
Only dependency and adjacency edges were included in these experiments.
Table~\ref{tb:pathsWeights} shows the results. 
Penalizing adjacency produces large gains; a harsh penalty is particularly helpful with fewer paths.
These results support the hypothesis that dependency edges are usually more meaningful for relation extraction than word adjacency. Therefore, if adjacency edges get the same weights, they might cause some dependency sub-paths drop out of the top $K$ paths, thus degrading performance. 
When the path number increases, there is a consistent and substantial increase in accuracy, which demonstrates the advantage of allowing adjacency edges to interleave with dependency ones. This presumably helps address syntactic parsing errors, among other things.
The importance of adjacency weights decreases with more paths, but it is still significantly better to penalize adjacency edges.

In the experiments mentioned above, cross-sentence extraction was conducted using minimal-span  candidates only. 
We expected that this would provide a reasonable safeguard to filter out many unlikely candidates.
As empirical validation, we also conducted experiments on cross-sentence extraction without the minimal-span restriction, using the base model.
Test accuracy dropped sharply from 81.7\% to 79.1\% (not shown in the table).

\begin{table}[t]
    \centering
    \begin{tabular}{lrr}
    \toprule
      \textbf{Relations} & \textbf{Single-Sent.} & \textbf{Cross-Sent.} \\
    \midrule
      Candidates &   169,168 &   332,969 \\
      $p\ge0.5$  &    32,028 &    64,828 \\
      $p\ge0.9$  &    17,349 &    32,775 \\
    \midrule
      GDKD & \multicolumn{2}{c}{162} \\
    \bottomrule
    \end{tabular}
    \caption{
        Unique drug-gene interactions extracted from PubMed Central articles, compared to the manually curated Gene Drug Knowledge Database (GDKD) used for distant supervision. $p$ signifies the output probability. GDKD contains 341 relations, but only 162 have specific drug references usable as distant supervision.
    }
    \label{tb:recall}
\end{table}

\begin{table}[t]
    \centering
    \begin{tabular}{lrr}
    \toprule
       & \textbf{Gene} & \textbf{Drug} \\
    \midrule
      GDKD & 140 & 80 \\
    \midrule
      Single-Sent. ($p\ge 0.9$) & 4036 & 311 \\
      Single-Sent. ($p\ge 0.5$) & 6189 & 347 \\
    \midrule
      Cross-Sent. ($p\ge 0.9$) & 5580 & 338 \\
      Cross-Sent. ($p\ge 0.5$) & 9470 & 373 \\
    \bottomrule
    \end{tabular}
    \caption{
        Numbers of unique genes and drugs in the Gene Drug Knowledge Database (GDKD) vs. DISCREX extractions.
    }
    \label{tb:gene-drug}
\end{table}

\subsection{PubMed-Scale Extraction}

Our ultimate goal is to extract knowledge from all available text.
First, we retrained DISCREX on all available distant-supervision data, not restricting to a subset of the folds as in the automatic evaluation.
We used the systems performing best on automatic evaluation, with features derived from 30 shortest paths between each entity pair, and minimal-span candidates within three sentences for cross-sentence extraction.
We then applied the learned extractors to all PubMed Central articles.
We grouped the extracted instances into unique drug-gene pairs.
The classifier output a probability for each instance.
The maximum probability of instances in a group was assigned to the relation as a whole.
Table~\ref{tb:recall} shows the statistics of extracted relations by varying the probability threshold.
Cross-sentence extraction obtained far more unique relations compared to single-sentence extraction, improving absolute recall by 89-102\%.
Table~\ref{tb:gene-drug} compares the number of unique genes and drugs.
DISCREX extractions cover far more genes and drugs compared to GDKD, which bode well for applications in precision medicine.

\subsection{Manual Evaluation}

Automatic evaluation accuracies can be overly optimistic.
To assess the true precision of DISCREX, we also conducted manual evaluation on extracted relations.
Based on the automatic evaluation, the accuracy is similar for single-sentence and cross-sentence extraction. So we focused on the latter.
We randomly sampled extracted relation instances and asked two researchers knowledgeable in precision medicine to evaluate their correctness.
For each instance, the annotators were provided with the provenance sentences where the drug-gene pair were highlighted.
The annotators assessed in each case whether some relation was mentioned for the given pair.

A total of 450 instances were judged: 150 were sampled randomly from all candidates (random baseline), 150 from the set of instances with probability no less than 0.5, and 150 with probability no less than 0.9.
From each set, we randomly selected 50 relations for review by both annotators.
The two annotators agreed on 133 of 150.
After review, all disagreements were resolved, and each annotator judged an additional set of 50 relation instances, this time without overlap.

Table~\ref{tb:manualEval} showed the sample precision and percentage of errors due to entity linking vs. relation extraction. 
With either classification threshold, cross-sentence extraction clearly outperformed the random baseline by a wide margin.
Not surprisingly, the higher threshold of 0.9 led to higher precision.
Interestingly, a significant portion of errors stems from mistakes in entity linking, as has been observed in prior work \cite{poon&al15}. 
Improved entity linking, either alone or joint with relation extraction, is an important future direction.

Based on these estimates, DISCREX extracted about 37,000 correct unique interactions at the threshold of 0.5, and about 20,000 at the threshold of 0.9. In both cases, it expanded the Gene Drug Knowledge Base by two orders of magnitude.

We also performed manual evaluation in the single-sentence setting.
As in the automatic evaluation, single-sentence precisions are similar though slightly higher at all thresholds.
This suggests that the candidate set is cleaner and the resulting predictions are more accurate.
However, the resulting recall is substantially lower, dropping by 46\% at a threshold of 0.5, and by 40\% at a threshold of 0.9.

\begin{table}[t]
    \centering
    \begin{tabular}{lrrr}
    \toprule
      & \textbf{Prec.} & \textbf{Entity Err.} & \textbf{Relation Err.} \\
    \midrule
    \multicolumn{4}{l}{\emph{Single-sentence extractions}} \\
      $\,$Random      & 31 & 52 & 17 \\
      $\,$ $p \ge 0.5$ & 61 & 25 & 15 \\
      $\,$ $p \ge 0.9$ & \bf{71} & 13 & 15 \\
    \midrule
    \multicolumn{4}{l}{\emph{Cross-sentence extractions}} \\
      $\,$Random      & 23 & 50 & 27 \\
      $\,$ $p \ge 0.5$ & 57 & 20 & 23 \\
      $\,$ $p \ge 0.9$ & \bf{61} & 13 & 26 \\
    \bottomrule
    \end{tabular}
    \caption{
      Sample precision and error percentage: comparison between the single sentence and cross-sentence extraction models at various thresholds.
      Single sentence extraction is slightly better at all thresholds, at the expense of substantially lower recall: a reduction of 40\% or more in terms of unique interactions.
    }
    \label{tb:manualEval}
\end{table}

\section{Conclusion}

We present the first approach for applying distant supervision to cross-sentence relation extraction, by adopting a document-level graph representation that incorporates both intra-sentential dependencies and inter-sentential relations such as adjacency and discourse relations.
We conducted both automatic and manual evaluation on extracting drug-gene interactions from biomedical literature.
With cross-sentence extraction, our DISCREX system doubled the yield of unique interactions, while maintaining the same accuracy.
Using distant supervision, DISCREX improved the coverage of the Gene Drug Knowledge Database (GDKD) by two orders of magnitude, without requiring annotated examples.

Future work includes: further exploration of features; improved integration with coreference and discourse parsing; combining distant supervision with active learning and crowd sourcing; evaluate the impact of extractions to precision medicine; applications to other domains.

\bibliographystyle{eacl2017}
\bibliography{refs}

\end{document}